\documentclass{article}
\usepackage{ijcai26}

\usepackage{times}
\usepackage{soul}
\usepackage{url}
\usepackage[hidelinks]{hyperref}
\usepackage[utf8]{inputenc}
\usepackage[small]{caption}
\usepackage{graphicx}
\usepackage{amsmath}
\usepackage{amsthm}
\usepackage{booktabs}
\usepackage{algorithm}
\usepackage{algorithmic}
\usepackage[switch]{lineno}
\usepackage{amssymb}  % 或者
\usepackage{amsfonts}
\newtheorem{definition}{Definition}
\usepackage{multirow}
\usepackage{booktabs}
\usepackage{colortbl}
\usepackage{xcolor}
\usepackage{subcaption}

\usepackage{array}
\usepackage{caption}
\usepackage[table]{xcolor}
\usepackage{tabularx}
\usepackage{ragged2e} % 提供更好的左对齐支持

\newcolumntype{L}[1]{>{\raggedright\arraybackslash}p{#1}}
\newcolumntype{C}[1]{>{\centering\arraybackslash}p{#1}}

\title{Efficient Time Series Clustering from Multiscale Reservoir Dynamics with Granular-Ball Anchoring Graph Optimization}

\author{
    Yifan Wang$^{1}$,~
    Lifeng Shen$^{1}${\thanks{Corresponding Author}},~
    Shuyin Xia$^{1}$,~
    Yi Wang$^{2}$\\   
    \affiliations
    $^{1}$ Chongqing Key Laboratory of Computational Intelligence, Key Laboratory of Cyberspace Big Data Intelligent Security, Ministry of Education, Sichuan-Chongqing Co-construction Key Laboratory of Digital Economy Intelligence and Key Laboratory of Big Data Intelligent Computing, College of Computer Science and Technology, Chongqing University of Posts and Telecommunications\\
    $^{2}$ Chongqing Ant Consumer Finance Co,. Ltd , Ant Group.\\
    \emails
    wangyifan421@foxmail.com,
    \href{mailto:shenlf@cqupt.edu.cn}{shenlf@cqupt.edu.cn},
    xiasy@cqupt.edu.cn, haonan.wy@myxiaojin.cn        
}

\begin{document}

\maketitle

\begin{abstract}

Time-series clustering remains challenging due to the inherent trade-off
between clustering effectiveness and computational efficiency.
Similarity-based methods often suffer from quadratic complexity caused by
pairwise distance computations, while deep learning–based approaches typically
rely on costly iterative training and a large number of trainable parameters. 
In this paper, we propose MSRGC-Net, an efficient time-series
clustering framework that integrates multiscale reservoir computing,
granular-ball-based anchoring graph construction, and consensus learning.
MSRGC-Net adopts a training-free reservoir computing paradigm to extract
multiscale temporal representations from raw time series without
backpropagation, significantly reducing computational overhead.
To capture the intrinsic structure of the resulting representations, granular-ball
computing is employed to adaptively model data distributions via
density-consistent regions, yielding compact and robust anchor graph representations.
Furthermore, a consensus-based anchoring graph optimization strategy is
introduced to effectively align multiscale reservoir representations and
integrate complementary information across temporal scales. 
Extensive experiments on widely used univariate and multivariate benchmark
datasets demonstrate that MSRGC-Net consistently outperforms state-of-the-art
methods in clustering performance while maintaining superior computational
efficiency.

\end{abstract}

\section{Introduction}

Massive time-series data
% \cite{1,2}
are ubiquitous in both scientific research and
real-world applications, including health monitoring~\cite{3}, human activity
recognition~\cite{ma2017walking}, speech analysis~\cite{5} and environmental science~\cite{6}.
As a fundamental unsupervised learning task, time-series clustering aims to
discover intrinsic temporal structures from unlabeled sequential data and plays
a crucial role in large-scale data analysis.

Early time-series clustering methods typically combine classical clustering
algorithms, such as k-means~\cite{18} and k-Shape~\cite{k-shape}, with distance
measures defined directly on raw time series, including Euclidean distance~\cite{16}
and Dynamic Time Warping (DTW)~\cite{17}.
Although conceptually simple, these distance-based approaches suffer from two
well-known limitations: they are sensitive to noise and complex temporal
variations, and they incur quadratic computational complexity due to repeated
pairwise distance calculations, which severely limits their scalability.

To alleviate these issues, feature-based clustering methods~\cite{22} first
transform time series into compact representations before clustering.
While improving efficiency and robustness, their performance strongly depends on
handcrafted features and domain expertise, and important temporal information
may be lost during feature extraction.
Moreover, deep learning–based approaches~\cite{26,27} have demonstrated
powerful representation learning capability.
However, their reliance on iterative training, extensive hyperparameter tuning,
and large computational budgets poses significant challenges for large-scale and
resource-constrained time-series clustering scenarios.

These limitations highlight a fundamental challenge in time-series clustering:
\emph{how to learn expressive temporal representations while maintaining high
computational efficiency}.
Addressing this challenge requires rethinking both representation learning and
clustering strategies.
In particular, two key issues remain largely unresolved.

First, temporal representations learned under different model configurations or
scales often capture complementary dynamic characteristics.
Although exploiting such diversity is beneficial for clustering, existing
methods typically rely on repeated training or complex ensembles, leading to
substantial computational overhead.
How can multiscale and diverse representations be generated efficiently without
costly training procedures?

Second, most existing methods apply k-means–based or spectral clustering directly
to learned representations, which assume simple cluster geometries and scale
poorly with dataset size.
How can clustering be redesigned to improve robustness and scalability while
avoiding expensive point-wise affinity modeling?

In this paper, we propose \textbf{MSRGC-Net} (\underline{M}ulti-\underline{S}cale
\underline{R}eservoir \underline{G}ranular-ball \underline{C}onsensus \underline{Net}work),
an efficient time-series clustering framework that integrates multi-scale
reservoir computing, granular-ball-based anchoring graph construction, and
consensus learning to address these challenges. 
MSRGC-Net adopts a training-free reservoir computing paradigm to extract
multiscale temporal representations using a set of fixed reservoirs with
different spectral radii, enabling efficient representation learning without
backpropagation.
To capture the intrinsic structure of the resulting reservoir representations,
granular-ball computing is employed to model data distributions via compact,
density-consistent regions, yielding a set of representative anchors.
Furthermore, a consensus-based anchoring graph optimization strategy is
introduced to align and integrate multiscale reservoir representations within a
unified clustering framework.
By combining training-free multiscale encoding, region-level abstraction, and
anchor-based consensus graph modeling, MSRGC-Net effectively balances
representation expressiveness and computational efficiency, making it well
suited for large-scale time-series clustering.

\vspace{0.1cm}
\noindent Main contributions are summarized as follows

i) We propose a training-free multiscale reservoir encoding mechanism that captures complementary temporal dynamics with different spectral radii.

ii) We introduce a granular-ball–based anchoring representation that compactly models the reservoir state distributions and enables efficient clustering.

iii) We formulate time-series clustering as a consensus-based anchoring graph optimization problem, providing a unified and lightweight framework
    for integrating multiscale dynamics.

\section{Related Work}

\subsection{Time-Series Clustering}

Time-series clustering aims to group unlabeled time series by
exploiting their intrinsic dynamics.
Early methods are mainly relying on similarity measures defined
directly on time series, such as DTW-based clustering~\cite{17} and
k-Shape~\cite{k-shape}.
Although effective in capturing local temporal patterns, these approaches
require pairwise distance computations and thus suffer from quadratic
complexity, limiting their scalability.

To improve efficiency, representation-based methods adopt a
``transform-then-cluster'' paradigm.
Representative approaches include statistical feature extraction~\cite{29},
kernel-based models such as TCK~\cite{TCK}, and feature fusion frameworks like
Time2Feat~\cite{time2feat}.
While reducing computational cost, these methods rely on handcrafted or
predefined representations and may fail to capture complex nonlinear dynamics.

More recently, deep learning–based methods learn data-adaptive temporal
representations, including autoencoder-based approaches~\cite{27,yang2024toward} and joint
learning frameworks~\cite{26,28}.
Despite their strong expressive power, these methods require iterative training
with many parameters, resulting in high computational cost and limited
scalability.
In contrast, our work achieves expressive temporal representations without
iterative network training by exploiting the intrinsic dynamics of reservoir
computing and coupling them with a lightweight clustering optimization
strategy.

\begin{figure}[t!]
    \centering
    \centering  \includegraphics[width=0.35\textwidth, trim=0 0 0 0, clip]{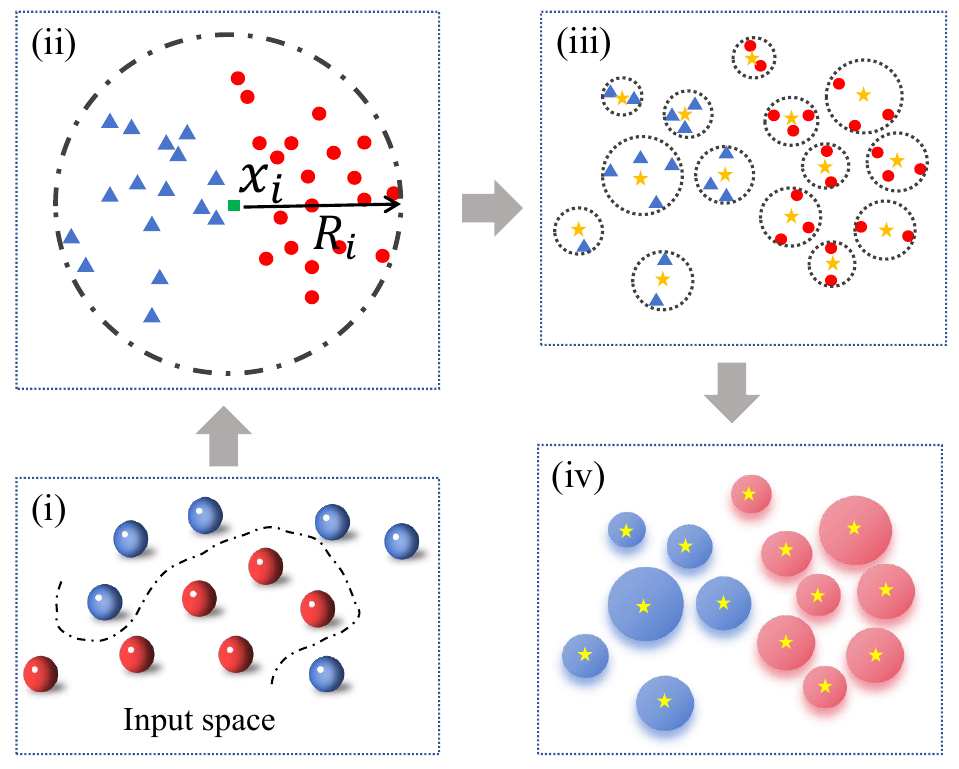}
    \caption{Example of using granular balls for anchoring.}
    \label{granular ball}% \vspace{-0.2cm}
\end{figure}

\begin{figure*}[t!]
    \centering
    \centering  \includegraphics[width=0.9\textwidth, trim=0 0 0 0, clip]{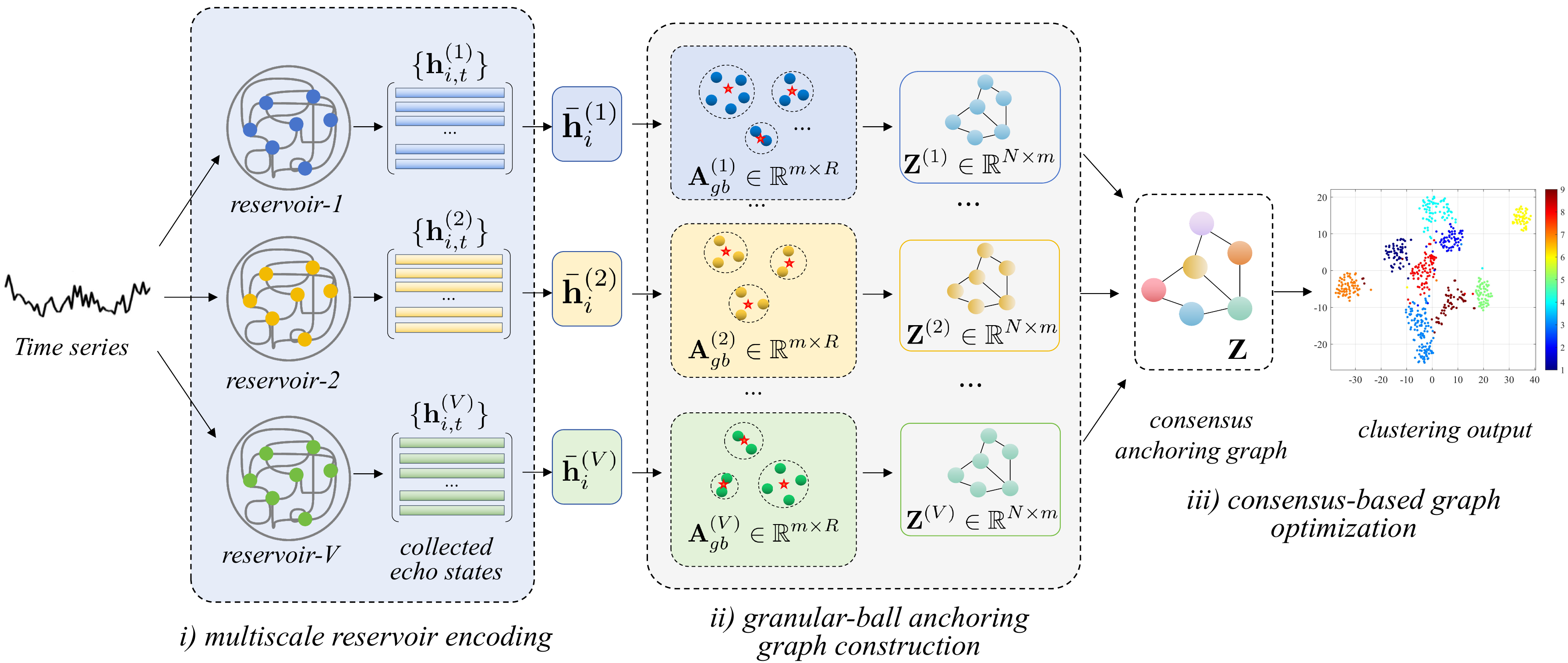}
    \caption{Illustration of the proposed MSRGC-Net.}
    \label{flow}
\end{figure*}

\subsection{Reservoir Computing for Time-Series}

Reservoir computing (RC), particularly Echo State Networks (ESNs)
\cite{jaeger2001echo}, provides an efficient framework for
temporal modeling by employing fixed recurrent dynamics and avoiding
backpropagation.
Owing to its training-free nature, RC has been widely applied to time-series
forecasting, classification, and anomaly detection
\cite{bianchi2015prediction,ullah2022intelligent}.
To enhance representation capacity, deep and multi-reservoir ESN architectures
have been explored to capture temporal dynamics at multiple scales
\cite{ma2020deepr}.
However, existing reservoir-based methods are mainly designed for supervised
tasks, and unsupervised approaches typically rely on a single reservoir and
apply conventional clustering directly to reservoir states, limiting their
ability to exploit complementary multiscale dynamics.

\subsection{Multi-View Graph Clustering}
Multi-view graph clustering \cite{34,35} is an important branch of multi-view clustering \cite{gao2023deep}.
Our work is also inspired by multi-view graph clustering, which fuses complementary
representations via graph-based optimization.
Although anchor-based methods~\cite{wang2022highly} improve scalability, their extension to time-series data remains limited due to the difficulty of capturing multiscale temporal representations.

Granular-ball computing (GBC) \cite{xia2019granular} provides a density-aware, region-level abstraction that enables each anchor to represent a compact local region rather than an isolated data point, as illustrated in Figure \ref{granular ball}. By operating at the regional level, GBC captures intrinsic local structures and bridges fine-grained instance-level information with higher-level semantic organization—an effect that is difficult to achieve using purely point-based representations~\cite{xia2025gbct,wang2026structure,xia2022efficient}. Such region-based modeling is particularly beneficial for multiscale reservoir representations~\cite{shen2026finding,xia2025graph}, where structural patterns are distributed across different temporal resolutions. Motivated by this observation, we incorporate granular-ball-based anchoring into a consensus-based multi-view graph optimization framework, enabling scalable and structure-aware time-series clustering.

\section{Methodology}

In this section, we formally elaborate on the proposed MSRGC-Net
(\underline{M}ulti-\underline{S}cale \underline{R}eservoir
\underline{G}ranular-ball \underline{C}onsensus \underline{Net}work),
a multiscale reservoir network with granular-ball-based anchoring graph
optimization for efficient time-series clustering.

As illustrated in Figure~\ref{flow}, MSRGC-Net consists of three main stages:
i) multiscale reservoir encoding, which extracts training-free temporal
representations from raw time series;
ii) granular-ball anchoring graph construction, which builds a structured
and compact graph representation to capture the intrinsic data organization;
and 
iii) consensus-based graph optimization for clustering, which integrates
multi-scale reservoir representations through lightweight optimization.

\subsection{Multi-Scale Reservoir Encoding}

Echo State Networks (ESNs) are a representative realization of reservoir computing, where a fixed recurrent neural network projects low-dimensional input sequences into a high-dimensional dynamical state space.
Given a multivariate time series dataset $\mathcal{X} = \{\mathbf{x}_i\}_{i=1}^N$ ($\mathbf{x}_i \in \mathbb{R}^{T \times D}$), the reservoir state of the $i$-th sample at time step $t$, denoted by $\mathbf{h}_{i,t} \in \mathbb{R}^{R}$, is obtained by the following update equation: 
\begin{equation}
    \mathbf{h}_{i,t} = \tanh\!\left( \mathbf{W}_{\mathrm{in}} \mathbf{x}_{i,t} + \mathbf{W}_{\mathrm{r}} \mathbf{h}_{i,t-1} \right),
\label{eq:esn_update}
\end{equation}
where $\mathbf{W}_{\mathrm{in}}\in\mathbb{R}^{R\times D}$ and $\mathbf{W}_{\mathrm{r}}\in\mathbb{R}^{R\times R}$ are the input and recurrent weight matrices, and $R$ is the number of reservoir units. In practice, $\mathbf{h}_i(0)$ can be initialized as a zero vector (or a small random vector), and the nonlinear activation $\tanh(\cdot)$ provides a bounded and smooth state transition to capture nonlinear temporal dependencies.  These weights are randomly initialized and kept fixed during learning, enabling efficient and stable extraction of temporal dynamics without backpropagation.

A single ESN configuration is generally inadequate for modeling the diverse
temporal dependencies exhibited by complex time series.
The dynamical behavior and memory capacity of an ESN are largely determined
by the spectral radius $\rho$ of the recurrent weight matrix
$\mathbf{W}_{\mathrm{r}}$.
Specifically, smaller values of $\rho$ favor short-term transient dynamics,
whereas values approaching unity drive the reservoir toward the
``edge of chaos,'' enabling the preservation of long-term temporal
dependencies~\cite{ma2020deepr}.
Motivated by this observation, we construct $V$ independent reservoirs with
distinct spectral radii $\mathcal{P}=\{\rho_1,\ldots,\rho_V\}$ to capture
complementary temporal patterns at multiple scales, thereby forming a set
of multi-scale reservoir views .

For the $i$-th time series processed by the $v$-th reservoir, a sequence of
reservoir states $\{\mathbf{h}_{i,t}^{(v)}\}_{t=1}^{T}$ is generated according
to Equation~(\ref{eq:esn_update}).
To obtain a compact view-specific representation, temporal aggregation is
applied to the reservoir states, yielding
\begin{equation}
    \bar{\mathbf{h}}_i^{(v)} = \frac{1}{T} \sum_{t=1}^{T} \mathbf{h}_{i,t}^{(v)}
    \in \mathbb{R}^{R}.
\end{equation}
By stacking the aggregated representations of all $N$ samples, we form the
view-specific feature matrix
$\mathbf{H}^{(v)} =
[\bar{\mathbf{h}}_1^{(v)}, \bar{\mathbf{h}}_2^{(v)}, \ldots,
\bar{\mathbf{h}}_N^{(v)}]^\top \in \mathbb{R}^{N \times R}$,
which is subsequently used as the input to the granular-ball anchoring graph
construction.

\subsection{Granular-Ball Anchoring Graph}

Given multi-scale reservoir representations
$\{\mathbf{H}^{(v)}\}_{v=1}^{V}$, directly constructing a consensus graph over
all samples is computationally prohibitive.
To enable scalable modeling while preserving the intrinsic structure of the
data, we construct a granular-ball anchoring graph based on granular-ball
computing (GBC)~\cite{xia2019granular}.
Specifically, GBC transforms each view-specific feature matrix
$\mathbf{H}^{(v)}$ into a compact set of representative anchors, which serve as
structural proxies of the original data distribution \cite{jiang2025scalable} and form the basis of an
anchor-based graph representation \cite{Cheng2024}.

Intuitively, granular-ball computing can be regarded as a region-inducing operator that
groups point-wise representations into adaptive local regions, thereby
capturing density and geometric structure beyond individual samples.
Given $\mathbf{H}^{(v)}$, the operator iteratively partitions the feature space
and outputs a collection of granular regions, referred to as granular balls.
This design is inspired by the global-priority principle in cognitive
computing and the multi-granularity learning framework. 
Below, we introduce the key components of the granular-ball anchoring graph
construction.

\begin{definition}\textbf{Granular Ball.}
Given the representations $\mathbf{H}^{(v)}$ of all samples,
granular-ball computing produces a set of granular balls
$\mathbf{GB}=\{GB_k\}_{k=1}^{M}$.
Each granular ball $GB_k$ is characterized by a center $\mathbf{c}_k$ and a
radius $r_k$, defined as
\begin{align}
\mathbf{c}_{k} &= \frac{1}{|GB_{k}|}
\sum_{\bar{\mathbf{h}}_{i_k}\in GB_{k}} \bar{\mathbf{h}}_{i_k}, \\ 
r_{k} &= \max_{\bar{\mathbf{h}}_{i_k}\in GB_{k}}
\|\bar{\mathbf{h}}_{i_k}-\mathbf{c}_{k}\|_2. 
\end{align}
\end{definition}

\begin{definition}\textbf{Quality of Granular Ball.}\label{Quality of Granular Ball}
The structural quality of a granular ball $GB_k$ is measured by the
distribution measure (DM)~\cite{xie2024mgnr}, defined as
\begin{equation}
DM(GB_k)=\frac{1}{|GB_k|}
\sum_{\bar{\mathbf{h}}_{i_k}\in GB_k}
\|\bar{\mathbf{h}}_{i_k}-\mathbf{c}_k\|_2,
\end{equation}
where a smaller DM value indicates a more compact and homogeneous region.
\end{definition}

\begin{definition}\textbf{Granular-Ball Splitting Criterion.}\label{Granular-Ball Splitting Criterion}
To adaptively refine granular regions, each granular ball $GB$ is tentatively
split into two sub-balls $GB_1$ and $GB_2$ using 2-means clustering.
The split is accepted if the following weighted criterion is satisfied:
\begin{equation}
DM_w=\frac{|GB_1|}{|GB|}DM(GB_1)+\frac{|GB_2|}{|GB|}DM(GB_2).
\end{equation}
If $DM_w < DM(GB)$, the split is retained; otherwise, the splitting process
terminates.
This procedure is repeated until no further quality-improving splits are
possible.
\end{definition}

Applying the granular-ball operator yields $M$ granular balls, where $M$
controls the granularity of the induced regions and balances structural
fidelity with computational efficiency.
To construct a compact anchoring graph, we select the top $m$ core
granular-ball anchors ($m<M<N$).
Anchor selection is guided by the following score:
\begin{equation}
\label{eq:gb_score}
score(GB_i)=DM(GB_i)\cdot
\min_{j\in\mathcal{S}_{q-1}}
\|\mathbf{c}_i-\mathbf{c}_j\|_2,
\end{equation}
where $\mathcal{S}_{q-1}$ denotes the set of previously selected anchors.
This criterion jointly accounts for regional compactness and spatial
diversity, enabling the selected anchors to preserve the skeletal structure
of the data distribution. 
%More details of granular computing are provided in Appendix~\ref{appendix:GB}.

Given $m$ granular-ball anchors for each view, denoted as
$\{\mathbf{A}_{gb}^{(v)}\in\mathbb{R}^{m\times R}\}_{v=1}^{V}$,
where each anchor corresponds to the center of a selected granular ball,
we construct a view-specific granular-ball anchoring graph
$\mathbf{Z}\in\mathbb{R}^{N\times m}$ to obtain a sample-to-anchor
representation \cite{liu2026views}.
Unlike conventional point-to-point affinity graphs, this anchoring graph
models relationships between samples and density-consistent granular-ball
anchors, whose centers summarize local data regions and capture intrinsic
structural patterns while suppressing the influence of local noise.
By replacing sample-level interactions with anchor-level connections, the
proposed representation significantly reduces memory and computational
complexity, while providing a robust and structurally faithful foundation
for subsequent consensus learning.

\subsection{Consensus Anchoring Graph Learning}
Based on $\{\mathbf{Z}^{(v)}\}_{v=1}^V$ the view-specific granular-ball anchoring graphs constructed in the
previous step, we further learn a unified consensus representation $\mathbf{Z}$ that
integrates complementary information from multiple views.
To jointly model view-wise data reconstruction, cross-view alignment, and
global structural regularization, we formulate the following unified
optimization problem:
\begin{equation}
\begin{aligned}
    \min_{\mathbf{Z},\, \{\mathbf{Z}^{(v)}\},\, \{\alpha^{(v)}\}}~
    & \sum_{v=1}^V
    \left\| \mathbf{H}^{(v)} - \mathbf{Z}^{(v)} \mathbf{A}_{gb}^{(v)} \right\|_F^2 \\
    & + \gamma \sum_{v=1}^V (\alpha^{(v)})^r
    \left\| \mathbf{Z} - \mathbf{Z}^{(v)} \right\|_F^2
    + \lambda \|\mathbf{Z}\|_F^2 \\
    \text{s.t.}\quad
    & \mathbf{Z} \ge 0,\ \mathbf{Z}^{(v)} \ge 0,\ 
    \mathbf{Z}\mathbf{1} = \mathbf{1},\ 
    \mathbf{Z}^{(v)}\mathbf{1} = \mathbf{1},
\end{aligned}
\label{obj}
\end{equation}
where $\mathbf{Z}^{(v)} \in \mathbb{R}^{N \times m}$ denotes the
view-specific sample-to-anchor assignment matrix, and
$\mathbf{Z} \in \mathbb{R}^{N \times m}$ represents the global consensus
anchoring graph.
Intuitively, the first term enforces faithful reconstruction of view-specific reservoir
representations using granular-ball anchors.
The second term encourages consistency between each view-specific anchoring
graph and the global consensus graph, while the third term imposes global
regularization to improve robustness and prevent overfitting. 
The adaptive weights $\{\alpha^{(v)}\}_{v=1}^V$ automatically adjust the
contribution of each view according to its alignment quality with the
consensus representation.
Following~\cite{chen2023learnable}, we set $r=2$ to ensure stable weight
smoothing and convergence behavior.
The parameters $\gamma$ and $\lambda$ control the strength of cross-view
consistency and global regularization, respectively.
Note that the objective function in Equation~(\ref{obj}) is not jointly convex with respect
to all variables.
We therefore adopt an alternating optimization strategy, iteratively updating
one variable while fixing the others until convergence.

\paragraph{Update of $\mathbf{Z}^{(v)}$.}
Fixing $\mathbf{Z}$ and $\{\alpha^{(v)}\}$, the subproblem with respect to
$\mathbf{Z}^{(v)}$ reduces to a regularized least-squares problem.
Its closed-form solution is given by
\begin{equation}
\resizebox{.9\linewidth}{!}{$
    \mathbf{Z}^{(v)} =
    \left( \mathbf{H}^{(v)} \mathbf{A}_{gb}^{(v)\top}
    + \gamma (\alpha^{(v)})^r \mathbf{Z} \right)
    \left( \mathbf{A}_{gb}^{(v)} \mathbf{A}_{gb}^{(v)\top}
    + \gamma (\alpha^{(v)})^r \mathbf{I} \right)^{-1}.
$}
\end{equation}
Negative entries are truncated to zero to satisfy the non-negativity
constraint.

\paragraph{Update of $\mathbf{Z}$.}
Fixing $\{\mathbf{Z}^{(v)}\}$ and $\{\alpha^{(v)}\}$, global consensus
anchoring graph admits the closed-form update
\begin{equation}
    \mathbf{Z} =
    \frac{\gamma \sum_{v=1}^V (\alpha^{(v)})^r \mathbf{Z}^{(v)}}
    {\gamma \sum_{v=1}^V (\alpha^{(v)})^r + \lambda},
\end{equation}
which corresponds to a weighted aggregation of view-specific anchoring graphs,
with adaptive emphasis on more consistent views.

\paragraph{Update of $\boldsymbol{\alpha}$.}
Fixing $\mathbf{Z}$ and $\{\mathbf{Z}^{(v)}\}$, the adaptive view weights
are updated via the Lagrange multiplier method as
\begin{equation}
    \alpha^{(v)} =
    \frac{(q^{(v)})^{\frac{1}{1-r}}}
    {\sum_{v=1}^V (q^{(v)})^{\frac{1}{1-r}}},
\end{equation}
where
$q^{(v)} = \gamma \|\mathbf{Z} - \mathbf{Z}^{(v)}\|_F^2$
measures the alignment error of the $v$-th view with respect to the consensus
graph.
Views with smaller alignment errors are thus assigned larger weights.
%More detailed derivations and pseudo-code are provided in Appendix~\ref{objective function solving}.

After convergence, the learned consensus anchoring graph
$\mathbf{Z} \in \mathbb{R}^{N \times m}$ serves as a low-rank affinity
representation of our MSRGC-Net.
Final cluster assignments are obtained via spectral clustering by performing
singular value decomposition:
\begin{equation}
    \mathbf{Z} = \mathbf{U} \boldsymbol{\Sigma} \mathbf{V}^\top,
\end{equation}
where $\mathbf{U} \in \mathbb{R}^{N \times k}$ contains the top-$k$ left
singular vectors.
The final labels are produced by applying $k$-means to the rows of
$\mathbf{U}$.

% \vspace{0.3cm}
% \noindent \textbf{Complexity analysis:}
\subsection{Complexity Analysis}
The computational complexity of MSRGC-Net follows its three-stage framework.
In the multiscale reservoir encoding stage, $V$ training-free reservoirs perform
forward propagation over $T$ time steps, resulting in a complexity of
$\mathcal{O}(V N T R^2)$, where $R$ denotes the reservoir size.
In the granular-ball anchoring graph construction stage, $m$ representative
anchors are generated with approximately linear complexity $\mathcal{O}(N R)$.
In the consensus-based graph optimization stage, MSRGC-Net learns a
sample-to-anchor graph $\mathbf{Z}\in\mathbb{R}^{N\times m}$, and the dominant
cost comes from the spectral decomposition on $\mathbf{Z}$ with complexity
$\mathcal{O}(N m^2)$.

Since both the reservoir size $R$ and the number of anchors $m$ are typically
fixed and satisfy $m \ll N$, MSRGC-Net achieves an overall near-linear complexity
with respect to the number of samples $N$.
Moreover, by integrating training-free reservoir computing and granular-ball
region abstraction with an anchor-based consensus graph, MSRGC-Net avoids costly
iterative optimization and point-wise affinity modeling, thereby offering a
clear efficiency advantage over conventional quadratic graph-based clustering
methods and deep models.

\begin{table*}[t!]
  \centering

  \resizebox{0.94\textwidth}{!}{
  \setlength{\tabcolsep}{1.8mm}
    \begin{tabular}{c|ccc|ccc|ccc|ccc|ccc}
    \toprule
    \multirow{2}[4]{*}{\textbf{Models}} 
    & \multicolumn{3}{c|}{\textbf{CT}} 
    & \multicolumn{3}{c|}{\textbf{JV}} 
    & \multicolumn{3}{c|}{\textbf{BM}} 
    & \multicolumn{3}{c|}{\textbf{Cric}} 
    & \multicolumn{3}{c}{\textbf{SCP1}} \\
\cmidrule{2-16}
    & NMI & ARI & RI 
    & NMI & ARI & RI 
    & NMI & ARI & RI
    & NMI & ARI & RI 
    & NMI & ARI & RI \\
    \midrule
    TCK   
    & 0.589 & 0.158 & 0.811  
    & 0.446 & 0.268 & 0.676  
    & \underline{0.689} & 0.603 & 0.841   
    & 0.812 & 0.647 & 0.947  
    & 0.141 & 0.198 & 0.621  \\
& {\scriptsize\textcolor{gray}{(0.004)}} 
& {\scriptsize\textcolor{gray}{(0.003)}} 
& {\scriptsize\textcolor{gray}{(0.002)}} 
& {\scriptsize\textcolor{gray}{(0.005)}} 
& {\scriptsize\textcolor{gray}{(0.004)}} 
& {\scriptsize\textcolor{gray}{(0.006)}}  
& {\scriptsize\textcolor{gray}{(0.006)}} 
& {\scriptsize\textcolor{gray}{(0.005)}} 
& {\scriptsize\textcolor{gray}{(0.007)}}  
& {\scriptsize\textcolor{gray}{(0.002)}} 
& {\scriptsize\textcolor{gray}{(0.003)}} 
& {\scriptsize\textcolor{gray}{(0.001)}}  
& {\scriptsize\textcolor{gray}{(0.008)}} 
& {\scriptsize\textcolor{gray}{(0.007)}} 
& {\scriptsize\textcolor{gray}{(0.005)}} \\

    K-shape   
    & 0.322 & 0.107 & 0.684  
    & 0.143 & 0.064 & 0.778  
    & 0.598 & 0.543 & 0.830 
    & 0.508 & 0.276 & 0.835 
    & 0.073 & 0.085 & 0.542 \\

& {\scriptsize\textcolor{gray}{(0.012)}} & {\scriptsize\textcolor{gray}{(0.008)}} & {\scriptsize\textcolor{gray}{(0.006)}}  
& {\scriptsize\textcolor{gray}{(0.015)}} & {\scriptsize\textcolor{gray}{(0.009)}} & {\scriptsize\textcolor{gray}{(0.008)}} 
& {\scriptsize\textcolor{gray}{(0.018)}} & {\scriptsize\textcolor{gray}{(0.015)}} & {\scriptsize\textcolor{gray}{(0.010)}}  
& {\scriptsize\textcolor{gray}{(0.011)}} & {\scriptsize\textcolor{gray}{(0.009)}} & {\scriptsize\textcolor{gray}{(0.007)}} 
& {\scriptsize\textcolor{gray}{(0.020)}} & {\scriptsize\textcolor{gray}{(0.015)}} & {\scriptsize\textcolor{gray}{(0.012)}} \\

    Fc-shape   
    & 0.421 & 0.059 & 0.721  
    & 0.210 & 0.059 & 0.634  
    & 0.664 & \underline{0.607} & 0.787 
    & 0.804 & 0.517 & 0.897 
    & 0.132 & 0.130 & 0.565 \\

& {\scriptsize\textcolor{gray}{(0.011)}} & {\scriptsize\textcolor{gray}{(0.006)}} & {\scriptsize\textcolor{gray}{(0.005)}}  
& {\scriptsize\textcolor{gray}{(0.014)}} & {\scriptsize\textcolor{gray}{(0.008)}} & {\scriptsize\textcolor{gray}{(0.009)}} 
& {\scriptsize\textcolor{gray}{(0.015)}} & {\scriptsize\textcolor{gray}{(0.012)}} & {\scriptsize\textcolor{gray}{(0.007)}}  
& {\scriptsize\textcolor{gray}{(0.010)}} & {\scriptsize\textcolor{gray}{(0.008)}} & {\scriptsize\textcolor{gray}{(0.004)}}  
& {\scriptsize\textcolor{gray}{(0.019)}} & {\scriptsize\textcolor{gray}{(0.013)}} & {\scriptsize\textcolor{gray}{(0.010)}} \\

    Modular-RC  
    & 0.476 & 0.199 & 0.819 
    & 0.126 & 0.059 & 0.656  
    & 0.577 & 0.325 & 0.620   
    & 0.469 & 0.221 & 0.754 
    & 0.214 & 0.148 & 0.575 \\

& {\scriptsize\textcolor{gray}{(0.003)}} & {\scriptsize\textcolor{gray}{(0.002)}} & {\scriptsize\textcolor{gray}{(0.001)}} 
& {\scriptsize\textcolor{gray}{(0.004)}} & {\scriptsize\textcolor{gray}{(0.003)}} & {\scriptsize\textcolor{gray}{(0.002)}} 
& {\scriptsize\textcolor{gray}{(0.005)}} & {\scriptsize\textcolor{gray}{(0.004)}} & {\scriptsize\textcolor{gray}{(0.003)}}  
& {\scriptsize\textcolor{gray}{(0.003)}} & {\scriptsize\textcolor{gray}{(0.002)}} & {\scriptsize\textcolor{gray}{(0.002)}}  
& {\scriptsize\textcolor{gray}{(0.006)}} & {\scriptsize\textcolor{gray}{(0.005)}} & {\scriptsize\textcolor{gray}{(0.004)}} \\

    DEC   
    & 0.665 & 0.406 & 0.926  
    & 0.611 & 0.477 & 0.888  
    & 0.529 & 0.391 & 0.744  
    & 0.590 & 0.323 & 0.895 
    & 0.116 & 0.134 & 0.567 \\

& {\scriptsize\textcolor{gray}{(0.032)}} 
& {\scriptsize\textcolor{gray}{(0.021)}} 
& {\scriptsize\textcolor{gray}{(0.014)}} 
& {\scriptsize\textcolor{gray}{(0.037)}} 
& {\scriptsize\textcolor{gray}{(0.045)}} 
& {\scriptsize\textcolor{gray}{(0.019)}}  
& {\scriptsize\textcolor{gray}{(0.030)}} 
& {\scriptsize\textcolor{gray}{(0.025)}} 
& {\scriptsize\textcolor{gray}{(0.015)}}  
& {\scriptsize\textcolor{gray}{(0.018)}} 
& {\scriptsize\textcolor{gray}{(0.027)}} 
& {\scriptsize\textcolor{gray}{(0.010)}}  
& {\scriptsize\textcolor{gray}{(0.051)}} 
& {\scriptsize\textcolor{gray}{(0.033)}} 
& {\scriptsize\textcolor{gray}{(0.024)}} \\

    GRAIL  
    & \underline{0.742} & \underline{0.608} & \underline{0.961} 
    & \underline{0.703} & \underline{0.581} & \underline{0.917}  
    & 0.538 & 0.309 & 0.673   
    & 0.774 & 0.699 & 0.832 
    & 0.203 & 0.112 & 0.577 \\

& {\scriptsize\textcolor{gray}{(0.008)}} 
& {\scriptsize\textcolor{gray}{(0.007)}} 
& {\scriptsize\textcolor{gray}{(0.003)}} 
& {\scriptsize\textcolor{gray}{(0.010)}} 
& {\scriptsize\textcolor{gray}{(0.009)}} 
& {\scriptsize\textcolor{gray}{(0.005)}}  
& {\scriptsize\textcolor{gray}{(0.012)}} 
& {\scriptsize\textcolor{gray}{(0.011)}} 
& {\scriptsize\textcolor{gray}{(0.014)}}  
& {\scriptsize\textcolor{gray}{(0.007)}} 
& {\scriptsize\textcolor{gray}{(0.006)}} 
& {\scriptsize\textcolor{gray}{(0.004)}}  
& {\scriptsize\textcolor{gray}{(0.021)}} 
& {\scriptsize\textcolor{gray}{(0.013)}} 
& {\scriptsize\textcolor{gray}{(0.009)}} \\

    Time2feat   
    & 0.686 & 0.410 & 0.905  
    & 0.610 & 0.417 & 0.869  
    & 0.654 & 0.600 & \textbf{0.873} 
    & \underline{0.921} & \underline{0.822} & \underline{0.966}  
    & 0.003 & 0.001 & 0.498 \\

& {\scriptsize\textcolor{gray}{(0.015)}} 
& {\scriptsize\textcolor{gray}{(0.017)}} 
& {\scriptsize\textcolor{gray}{(0.016)}} 
& {\scriptsize\textcolor{gray}{(0.018)}} 
& {\scriptsize\textcolor{gray}{(0.015)}} 
& {\scriptsize\textcolor{gray}{(0.009)}}  
& {\scriptsize\textcolor{gray}{(0.020)}} 
& {\scriptsize\textcolor{gray}{(0.019)}} 
& {\scriptsize\textcolor{gray}{(0.008)}}  
& {\scriptsize\textcolor{gray}{(0.010)}} 
& {\scriptsize\textcolor{gray}{(0.008)}} 
& {\scriptsize\textcolor{gray}{(0.000)}}  
& {\scriptsize\textcolor{gray}{(0.001)}} 
& {\scriptsize\textcolor{gray}{(0.001)}} 
& {\scriptsize\textcolor{gray}{(0.005)}} \\

    TimesURL   
    & 0.655 & 0.403 & 0.919  
    & 0.135 & 0.028 & 0.554  
    & 0.515 & 0.399 & 0.712 
    & 0.415 & 0.344 & 0.703 
    & 0.126 & 0.148 & 0.574 \\

& {\scriptsize\textcolor{gray}{(0.018)}} & {\scriptsize\textcolor{gray}{(0.015)}} & {\scriptsize\textcolor{gray}{(0.007)}}  
& {\scriptsize\textcolor{gray}{(0.020)}} & {\scriptsize\textcolor{gray}{(0.008)}} & {\scriptsize\textcolor{gray}{(0.010)}} 
& {\scriptsize\textcolor{gray}{(0.022)}} & {\scriptsize\textcolor{gray}{(0.018)}} & {\scriptsize\textcolor{gray}{(0.012)}}  
& {\scriptsize\textcolor{gray}{(0.015)}} & {\scriptsize\textcolor{gray}{(0.012)}} & {\scriptsize\textcolor{gray}{(0.009)}}  
& {\scriptsize\textcolor{gray}{(0.025)}} & {\scriptsize\textcolor{gray}{(0.020)}} & {\scriptsize\textcolor{gray}{(0.016)}} \\

    TFMCC   
    & 0.647 & 0.409 & 0.922  
    & 0.188 & 0.067 & 0.698  
    & 0.625 & 0.468 & 0.766   
    & 0.506 & 0.243 & 0.853 
    & 0.167 & 0.211 & 0.605 \\

& {\scriptsize\textcolor{gray}{(0.016)}} & {\scriptsize\textcolor{gray}{(0.014)}} & {\scriptsize\textcolor{gray}{(0.006)}}  
& {\scriptsize\textcolor{gray}{(0.018)}} & {\scriptsize\textcolor{gray}{(0.009)}} & {\scriptsize\textcolor{gray}{(0.011)}} 
& {\scriptsize\textcolor{gray}{(0.020)}} & {\scriptsize\textcolor{gray}{(0.015)}} & {\scriptsize\textcolor{gray}{(0.010)}}  
& {\scriptsize\textcolor{gray}{(0.014)}} & {\scriptsize\textcolor{gray}{(0.011)}} & {\scriptsize\textcolor{gray}{(0.008)}}  
& {\scriptsize\textcolor{gray}{(0.028)}} & {\scriptsize\textcolor{gray}{(0.022)}} & {\scriptsize\textcolor{gray}{(0.014)}} \\

    GB-SMKKM   
    & 0.559 & 0.338 & 0.931  
    & 0.234 & 0.112 & 0.807  
    & \textbf{0.692} & 0.558 & 0.840   
    & 0.895 & 0.761 & \underline{0.966} 
    & 0.207 & \underline{0.263} & \underline{0.631} \\

& {\scriptsize\textcolor{gray}{(0.005)}} & {\scriptsize\textcolor{gray}{(0.004)}} & {\scriptsize\textcolor{gray}{(0.002)}}  
& {\scriptsize\textcolor{gray}{(0.006)}} & {\scriptsize\textcolor{gray}{(0.005)}} & {\scriptsize\textcolor{gray}{(0.004)}}  
& {\scriptsize\textcolor{gray}{(0.007)}} & {\scriptsize\textcolor{gray}{(0.006)}} & {\scriptsize\textcolor{gray}{(0.005)}}  
& {\scriptsize\textcolor{gray}{(0.004)}} & {\scriptsize\textcolor{gray}{(0.004)}} & {\scriptsize\textcolor{gray}{(0.002)}}  
& {\scriptsize\textcolor{gray}{(0.008)}} & {\scriptsize\textcolor{gray}{(0.007)}} & {\scriptsize\textcolor{gray}{(0.005)}} \\

    MV-CAGAF   
    & 0.529 & 0.467 & 0.868  
    & 0.695 & 0.235 & \underline{0.917}  
    & 0.650 & 0.526 & 0.843   
    & 0.774 & 0.636 & 0.930 
    & \textbf{0.232} & 0.161 & 0.585 \\

& {\scriptsize\textcolor{gray}{(0.006)}} & {\scriptsize\textcolor{gray}{(0.005)}} & {\scriptsize\textcolor{gray}{(0.003)}}  
& {\scriptsize\textcolor{gray}{(0.008)}} & {\scriptsize\textcolor{gray}{(0.006)}} & {\scriptsize\textcolor{gray}{(0.004)}}  
& {\scriptsize\textcolor{gray}{(0.009)}} & {\scriptsize\textcolor{gray}{(0.008)}} & {\scriptsize\textcolor{gray}{(0.006)}}  
& {\scriptsize\textcolor{gray}{(0.005)}} & {\scriptsize\textcolor{gray}{(0.005)}} & {\scriptsize\textcolor{gray}{(0.003)}}  
& {\scriptsize\textcolor{gray}{(0.010)}} & {\scriptsize\textcolor{gray}{(0.009)}} & {\scriptsize\textcolor{gray}{(0.007)}} \\

    \midrule
    \textbf{MSRGC-Net}  
    & \textbf{0.789}& \textbf{0.645} & \textbf{0.964}  
    & \textbf{0.781} & \textbf{0.750} & \textbf{0.948}  
    & 0.678 & \textbf{0.615} & \underline{0.857}  
    & \textbf{0.929} & \textbf{0.884} & \textbf{0.983}  
    & \underline{0.225} & \textbf{0.270} & \textbf{0.635} \\
    % & \textcolor{gray}{0.003}& \textcolor{gray}{0.006} & \textcolor{gray}{0.001} 
    % & 0.004 & 0.003 & 0.001  
    % & 0.001 & 0.005 & 0.004  
    % & 0.002 & 0.002 & 0.000  
    % & 0.006 & 0.006 & 0.004 \\
  & {\scriptsize\textcolor{gray}{(0.003)}} 
& {\scriptsize\textcolor{gray}{(0.006)}} 
& {\scriptsize\textcolor{gray}{(0.001)}} 
& {\scriptsize\textcolor{gray}{(0.004)}} 
& {\scriptsize\textcolor{gray}{(0.003)}} 
& {\scriptsize\textcolor{gray}{(0.001)}}  
& {\scriptsize\textcolor{gray}{(0.001)}} 
& {\scriptsize\textcolor{gray}{(0.005)}} 
& {\scriptsize\textcolor{gray}{(0.004)}}  
& {\scriptsize\textcolor{gray}{(0.002)}} 
& {\scriptsize\textcolor{gray}{(0.002)}} 
& {\scriptsize\textcolor{gray}{(0.000)}}  
& {\scriptsize\textcolor{gray}{(0.006)}} 
& {\scriptsize\textcolor{gray}{(0.006)}} 
& {\scriptsize\textcolor{gray}{(0.004)}} \\

    \bottomrule
    \end{tabular}
    }  

\raggedright
% \parbox{\textwidth}{\scriptsize \textit{Note.} CT--CharacterTrajectories; JV--JapaneseVowels; BM--BasicMotions; Cric--Cricket; SCP1--SelfRegulationSCP1.}
  \caption{Clustering results on multivariate time series datasets.
Abbreviations: CT - CharacterTrajectories; JV - JapaneseVowels; BM - BasicMotions;
Cric - Cricket; SCP1 - SelfRegulationSCP1.}
    \label{multivariate}
\end{table*}

\section{Experiments}

\begin{table*}[t!]
\centering
% 将宽度调整为 \textwidth 以容纳更多列
\resizebox{0.95\textwidth}{!}{
\setlength{\tabcolsep}{3pt} % 稍微减小列间距以防表格过宽
% 注意这里的 |ccccc|ccccc，中间的竖线实现了你的要求
\begin{tabular}{l|ccc|ccccc|ccccc}
\toprule
\multirow{2}{*}{\textbf{Variants}} & \textbf{Multiscale} & \textbf{Granular-ball} & \textbf{Explicit} & \multicolumn{5}{c|}{\textbf{Multivariate Datasets}} & \multicolumn{5}{c}{\textbf{Univariate Datasets}} \\
 & \textbf{Reservoirs} & \textbf{Anchoring} & \textbf{Optim.} & \textbf{CT} & \textbf{JV} & \textbf{BM} & \textbf{Cric} & \textbf{SCP1} & \textbf{BC} & \textbf{Crop} & \textbf{EPG-R} & \textbf{EPG-S} & \textbf{Wafer}\\
\midrule
1. w/o Multiscale & $\times$ & $\checkmark$ & $\checkmark$ 
& 0.943 & 0.785 & 0.844 & 0.944 & 0.581 
& 0.738 & 0.926 & 0.993 & 0.997 & 0.559 \\ % 请填入新数据

2. w/o Granular-ball & $\checkmark$ & $\times$ & $\checkmark$ 
& 0.953 & 0.920 & 0.832 & 0.970 & 0.579 
& 0.740 & 0.922 & \textbf{1.000} & 0.984 & 0.561 \\ % 请填入新数据

3. w/o Optimization & $\checkmark$ & $\checkmark$ & $\times$  
& 0.952 & 0.936 & 0.811 & 0.966 & 0.559 
& 0.713 & 0.894 & 0.962 & 0.958 & 0.490 \\ % 请填入新数据

\midrule
\textbf{MSRGC-Net (Ours)} & $\checkmark$ & $\checkmark$ & $\checkmark$ 
& \textbf{0.964} & \textbf{0.948} & \textbf{0.857} & \textbf{0.983} & \textbf{0.635} 
& \textbf{0.759} & \textbf{0.941} & \textbf{1.000} & \textbf{1.000} & \textbf{0.575} \\ % 请填入新数据
\bottomrule
\end{tabular}
}
\caption{Ablation study results evaluated using the RI metric.}\label{tab:ablation_RI}
\end{table*}

% \subsection{
% Setup
% }
\noindent\textbf{Dataset.}
We evaluate the proposed MSRGC-Net on ten benchmark time-series datasets drawn from the UCR and
UEA archives~\cite{dau2019ucr,bagnall2018uea}, covering a wide range of application
domains, including human activity recognition, healthcare, industrial manufacturing, speech processing, etc.
The datasets span diverse sequence lengths and dataset sizes.
%Detailed statistics are provided in Appendix~\ref{appendix:data}.

\vspace{0.1cm}
\noindent\textbf{Evaluation metrics.}
Clustering performance is evaluated using three standard metrics \cite{rand1971objective}: Normalized
Mutual Information (NMI), Adjusted Rand Index (ARI), and Rand Index (RI).
% \cite{rand1971objective}.
Together, these metrics provide complementary information-theoretic and
pairwise-consistency evaluations.
%Detailed definitions are given in Appendix~\ref{appendix:Metrics}.

\vspace{0.1cm}
\noindent\textbf{Baselines.}
We compare MSRGC-Net with representative time-series clustering methods from three
categories.
\textit{(i) Raw data-based methods} operate directly on the original time
series, including k-Shape~\cite{k-shape}, Fuzzy-kShape~\cite{fuzzy-kshape}, and
TCK~\cite{TCK}.
\textit{(ii) Representation learning-based methods} perform clustering in a
learned feature space, covering both feature learning and deep clustering
approaches, such as Modular-RC~\cite{18}, GRAIL~\cite{GRAIL},
Time2Feat~\cite{time2feat}, DEC~\cite{DEC}, TimeSURL~\cite{timesURL}, and
TFMCC~\cite{TFMCC}.
\textit{(iii) Multi-view clustering methods} integrate information from multiple
views, including GB-SMKKM~\cite{GB-MKKM} and MV-CAGAF~\cite{MV-CAGAF}.
%Additional details for the above baselines are given in Appendix~\ref{appendix:methods}.

\vspace{0.1cm}
\noindent\textbf{Implementation details.}
A common set of reservoir hyperparameters was used across all datasets,
including reservoir size $R=400$, connectivity $\beta=0.25$, input scaling
$\omega=0.15$, noise level $\xi=0.001$, number of views $V=3$, and a spectral
radius determined by a predefined strategy.
For the consensus graph optimization, two regularization parameters were tuned
from small candidate sets via grid search, with the multi-view alignment
coefficient $\gamma=10^{3}$
and the structural constraint parameter $\lambda=10^{-1}$.
All experiments were repeated ten times with random initializations and
conducted on a 64-bit machine with an Intel i7-12700 CPU and 64~GB RAM.

\subsection{Main Results}

As shown in Table~\ref{multivariate}, MSRGC-Net achieves competitive and stable
performance on multivariate time series clustering tasks, yielding favorable
results compared with ten representative and state-of-the-art baseline methods
across NMI, ARI, and RI.
Across 5 multivariate datasets and 15 evaluation results, MSRGC-Net
obtains the best performance on 12 metrics and ranks second on two metrics,
indicating consistent behavior across different settings.

Improvements are more noticeable on datasets with higher channel
dimensions and longer sequences, where modeling cross-variable temporal
dependencies becomes increasingly important.
For example, on the JapaneseVowels (JV) dataset, which involves variable-length
and strongly coupled speech signals, MSRGC-Net achieves an ARI of 0.750, improving
upon the second-best method GRAIL by 16.9\%.
On the Cricket (Cric) dataset with ultra-long time series ($T=1197$), MSRGC-Net
maintains strong performance across all three metrics, with an average score of
0.932.
Similarly, on the CharacterTrajectories (CT) dataset containing 20 classes,
MSRGC-Net attains the highest NMI of 0.789, markedly outperforming the single-view
baseline rm-ESN (NMI = 0.476).
Compared with raw data-based methods, MSRGC-Net leverages reservoir-based feature
extraction to capture temporal dynamics across multiple variables.
Relative to deep learning approaches, MSRGC-Net attains comparable
or improved performance without iterative training.
Overall, these observations suggest that the proposed consensus learning
framework is effective for multivariate time series clustering.
%More univariate results are in Appendix~\ref{appendix: univariate}.

\begin{figure*}[t!]
    \centering
    \centering  \includegraphics[width=1\textwidth, trim=0 0 0 0, clip]{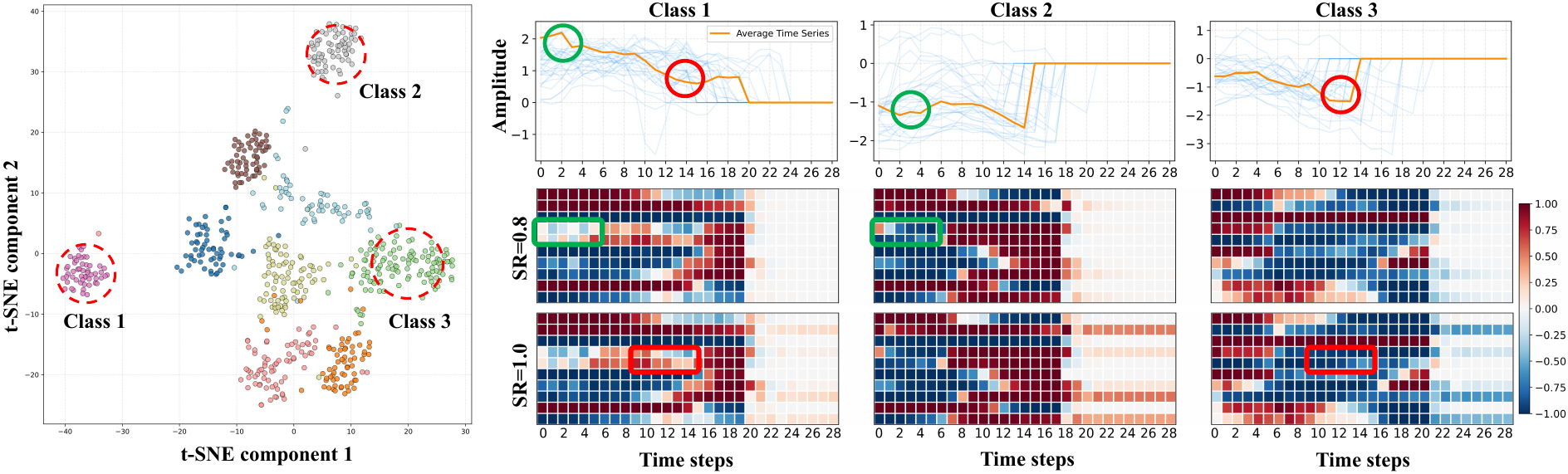}
    \caption{Visualization of complementary representations from multi-scale
reservoirs, where green and red annotations indicate characteristic dynamical
differences across clusters.}
    \label{representation capability}
\end{figure*}

\subsection{Ablation Studies}
To analyze the contribution of each component, we compare MSRGC-Net with three
ablated variants:
(1) \textbf{w/o Multi-scale}, where all reservoirs share a fixed spectral radius
of 0.9 to remove multi-scale dynamics;
(2) \textbf{w/o Granular-ball}, where adaptive granular-ball construction is
replaced by $k$-means clustering for anchor selection;
(3) \textbf{w/o Optimization}, where the optimization objective
(Eq.~\ref{obj}) is replaced by simple average fusion.

Table~\ref{tab:ablation_RI} summarizes the ablation results in terms of the RI
metric.
On the multivariate datasets, the full MSRGC-Net achieves the highest average RI
(0.883), outperforming the w/o Multi-scale, w/o Granular-ball, and w/o
Optimization variants by clear margins.
Similar trends are observed on univariate datasets, where the full model
consistently achieves the best or tied-best performance.
This demonstrates that all components contribute to the final
clustering performance, with the multi-scale mechanism playing a dominant role,
while granular-ball anchors and optimization-based fusion further enhance the
results.

\subsection{Visualization Analysis}
Figure~\ref{representation capability} visualizes the learned representations of
the proposed model on the JV multimodal dataset.
On the left, t-SNE projects the samples into a low-dimensional space, where three
representative clusters (Class~1/2/3) are identified for further analysis,
showing a clear separation in the learned representation space. 
The top-right panel illustrates the trajectories of samples within each cluster
(blue: individual samples; orange: cluster mean). 
The bottom-right panel presents ESN state heatmaps with spectral radii
SR$=0.8$ and SR$=1.0$, which provide a more fine-grained view of the underlying
dynamical representations.

Green markers highlight early-stage differences between Class~1 and Class~2, with
distinct temporal dynamics.
Red markers indicate mid-stage differences between Class~1 and Class~3, showing
clear dynamical changes. 
A vertical comparison across spectral radii reveals complementary temporal
patterns: smaller SR emphasizes local transient variations, whereas larger
SR captures longer-term dependencies.
This suggests that the multi-scale design enables the model to
encode diverse temporal characteristics, thereby enhancing cluster
separability in multimodal time-series data.

\subsection{Efficiency--Effectiveness Analysis}
\begin{figure}[t!]
    \centering
    \centering  \includegraphics[width=0.45\textwidth, trim=0 0 0 0, clip]{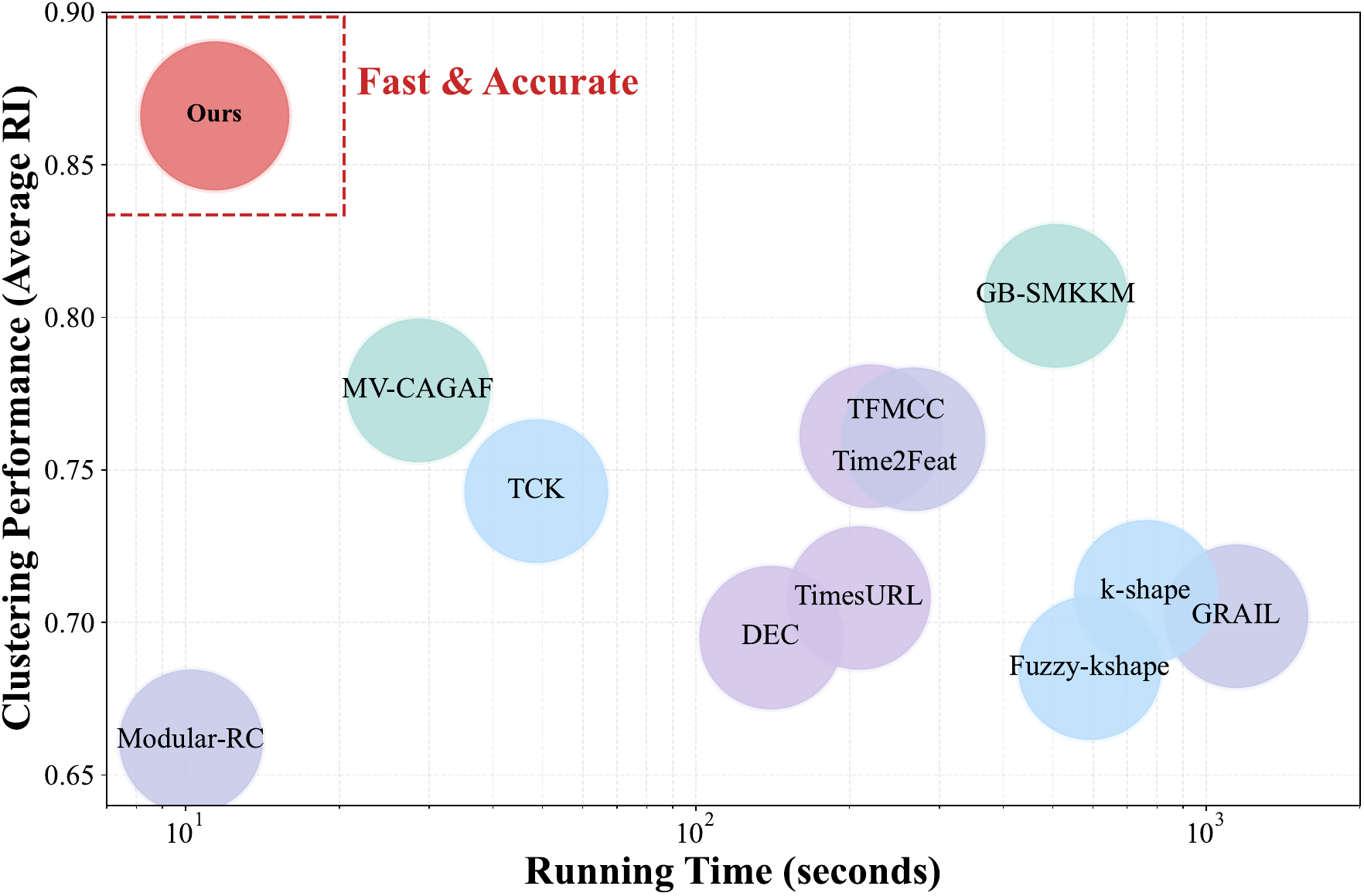}
    \caption{Running time versus clustering effectiveness.}
    \label{fig:Efficiency}\vspace{-0.2cm}
\end{figure}

Figure~\ref{fig:Efficiency} presents an efficiency-effectiveness comparison of all methods across ten datasets spanning both univariate and multivariate settings. The y-axis reports the average RI, while the x-axis shows the average running time on a logarithmic scale. Overall, our method lies in the upper-left region of the plot and on the Pareto
frontier, indicating a favorable trade-off between clustering performance and
computational cost.
It achieves high clustering accuracy with a runtime on the order of tens of
seconds. 
In contrast, several baseline methods incur substantially higher computational
costs without corresponding performance gains.
Compared with the second-best method, GB-SMKKM, our approach achieves superior
performance with roughly an order-of-magnitude lower runtime.
Moreover, the comparison with rm-ESN highlights the benefit of multi-view
reservoir dynamics over single-view representations.

\begin{figure}[h!]
    \centering
    \centering  \includegraphics[width=0.47\textwidth, trim=0 0 0 0, clip]{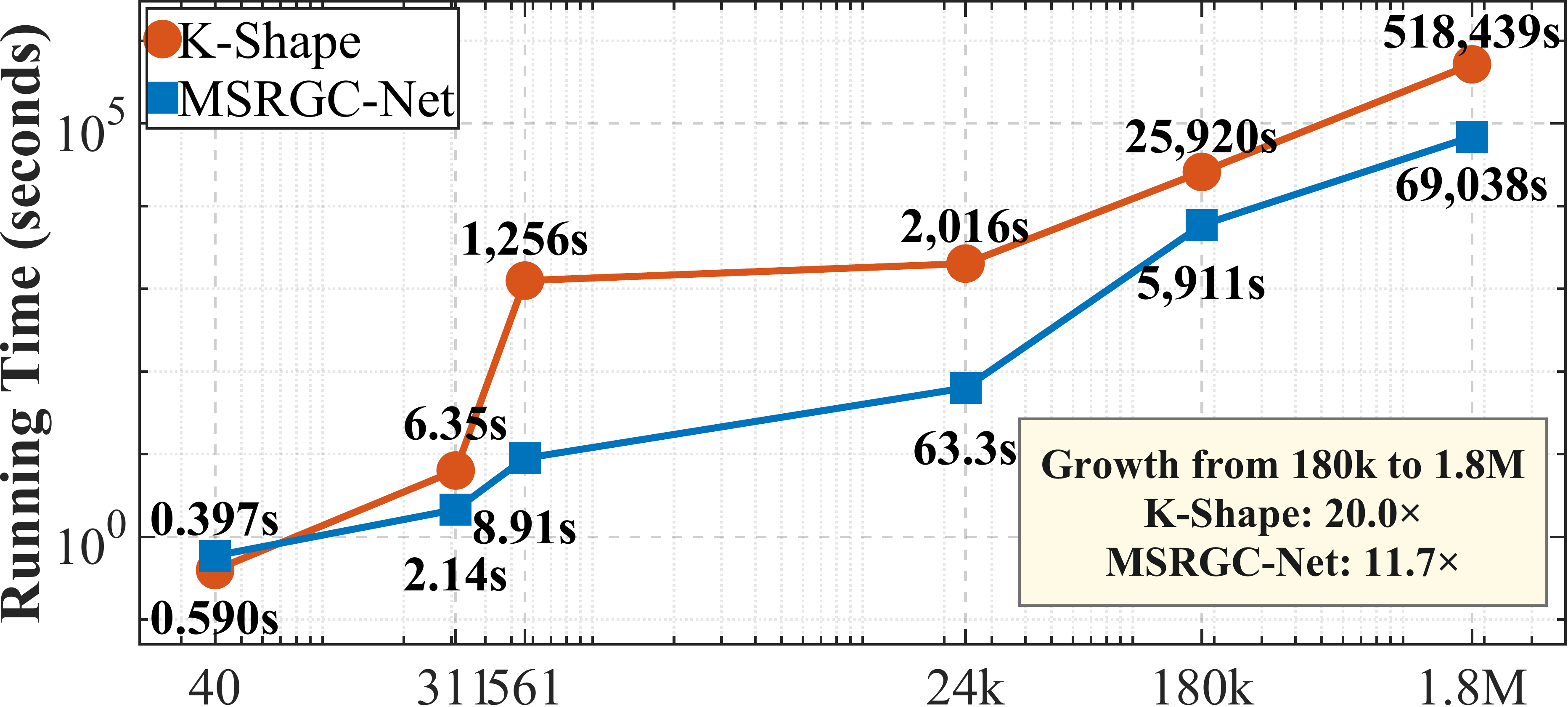}
    \caption{Scalability evaluation on million-scale samples.}
    \label{large-scale dataset}\vspace{-0.2cm}
\end{figure}

\noindent \textbf{Large-scale testing.} 
We further evaluate scalability on an extended Pedestrian dataset containing approximately 1.8 million samples. As shown in Figure~\ref{large-scale dataset}, the runtime increases near-linearly with the sample size and remains stable even at the million-sample scale. MSRGC-Net achieves an RI of 
0.947$\pm$0.002, outperforming k-shape (0.882$\pm$0.004), indicating that improved scalability does not come at the expense of clustering quality. 
This empirical observation is consistent with the theoretical time complexity
\(
\mathcal{O}(VNTR^2 + NR + Nm^2),
\)
where 
\(m \ll N\), implying near-linear growth with respect to 
$N$.

\section{Conclusion}
We proposed MSRGC-Net, a multi-scale reservoir-based graph clustering framework
for efficient time-series clustering.
By employing training-free reservoirs with diverse spectral radii, the model captures
complementary temporal dynamics ranging from local transient patterns to
long-term dependencies.
These multi-scale representations are integrated via granular-ball-based anchor
construction and a consensus graph optimization scheme, enabling effective
fusion of multi-view temporal information.
Extensive experiments on both univariate and multivariate datasets demonstrate
that MSRGC-Net consistently outperforms state-of-the-art methods.
Ablation and visualization analyses further validate the contribution of each
component and reveal how the multi-scale design enhances cluster separability.
Moreover, efficiency--effectiveness analysis shows that MSRGC-Net achieves a favorable balance between clustering accuracy and computational cost.

\section*{Acknowledgments}
This research was supported by the Chongqing Graduate Research Innovation Program CYB25259 and the National Natural Science Foundation of China under Grant Nos. 62221005, 62450043, 62222601, and 62176033.

\bibliographystyle{named}
\bibliography{ijcai26}

\end{document}